%% file: main.tex
\documentclass[10pt,twocolumn,letterpaper]{article}

\usepackage[pagenumbers]{cvpr} 

\input{preamble}
\definecolor{cvprblue}{rgb}{0.21,0.49,0.74}
\usepackage[pagebackref,breaklinks,colorlinks,allcolors=cvprblue]{hyperref}

\usepackage{hyperref}
\usepackage{url}
\usepackage{booktabs,graphicx}

\def\paperID{*****} 
\def\confName{CVPR}
\def\confYear{2027}

\title{SkeleWAM: Skeleton World-Action Modeling for Efficient Robotic Manipulation}

\author{Juyi Sheng\\
Peking University\\
{\tt\small Logss2024@stu.pku.edu.cn}
\and
Hua Wang\\
Peking University\\
{\tt\small huawang@stu.pku.edu.cn}
\and
Mengyuan Liu\\
Peking University\\
{\tt\small liumengyuan@pku.edu.cn}
}

\begin{document}
\maketitle
\input{sec/0_abstract}    
\input{sec/1_intro}
\input{sec/2_related.tex}
\input{sec/3_method.tex}
\input{sec/4_experiment.tex}

\input{sec/5_conclusion.tex}

{
    \small
    \bibliographystyle{ieeenat_fullname}
    \bibliography{main}
}


\end{document}

%% file: sec/0_abstract.tex
\begin{abstract}
World action models (WAMs) combine robot action generation with future state prediction. Existing WAMs typically predict videos or learned visual latents, which represent interaction geometry only implicitly and may retain appearance information unrelated to control. We introduce \textbf{SkeleWAM}, a compact WAM that represents a manipulation scene as a sparse 3D skeleton composed of robot joints, object centers, and interaction points. Constructed online from current RGB-D observations and robot proprioception, the skeleton provides a unified geometric state for action generation and future skeleton prediction. Future skeleton prediction provides additional geometric supervision for action learning without requiring visual reconstruction. At inference, SkeleWAM generates actions directly from the current skeleton and language instruction, while Medoid Action Consensus (MAC) serves as an auxiliary consensus strategy for stochastic action samples. On LIBERO-Plus, SkeleWAM achieves an overall success rate of 85.9\% with 57.1M parameters, outperforming Cosmos-Policy by 3.7 percentage points. These results demonstrate that sparse 3D robot--object structure provides an effective state space for robust and parameter-efficient world action learning. The project is available at \url{https://skelewam-project.github.io/}.
\end{abstract}

%% file: sec/1_intro.tex
\section{Introduction}
\label{sec:introduction}

Learning-based robotic manipulation has rapidly evolved from task-specific visuomotor policies to general-purpose robot foundation models. ACT learns action chunks from visual observations, while Diffusion Policy and DP3 model expressive action distributions using 2D and 3D visual representations, respectively \citep{act,chi2023diffusionpolicy,Ze2024DP3}. As model pretraining and robot datasets continue to scale, generalist policies and vision-language-action (VLA) models, including RT-2, Octo, OpenVLA, $\pi_0$, and $\pi_{0.5}$, have demonstrated increasingly broad capabilities across diverse manipulation tasks and environments \citep{rt2,octo_2023,openvla,pi0,intelligence2025pi05visionlanguageactionmodelopenworld}. World action models (WAMs) jointly model robot action generation and future state prediction, where the latter provides additional supervision for learning how manipulation scenes evolve under robot actions \citep{dreamzero,lingbotva,fastwam,kim2026cosmos}. Many existing WAMs represent future states as generated videos or learned visual latents \citep{dreamzero,lawam,being07}. Although latent prediction reduces the computational cost of explicit video generation, the information retained by visual latents is largely shaped by the underlying visual encoder and its pretraining objective. More importantly, compressing visual observations does not by itself isolate the geometric variables that govern robot--object interactions, such as their spatial configurations and relative motions. This raises a fundamental question: \emph{can a sparse representation of robot--object structure support effective world action learning while substantially reducing model complexity?}

Our key observation is that the aspects of a manipulation scene most relevant to control can be represented as a \emph{skeleton} consisting of a sparse set of 3D landmarks on the robot and manipulated objects. The robot is represented by the 3D positions of its joints, while each object is represented by its center and a set of interaction points connected to that center. The object center captures its overall spatial location, whereas the interaction points identify locations relevant to robot contact and manipulation. Together, these landmarks provide a compact and explicit representation of the robot configuration and object interaction geometry. We use this unified skeleton as the state representation and jointly learn robot action generation and future skeleton prediction without reconstructing visual appearance.

\begin{figure*}[t]
    \centering
    \includegraphics[width=1.0\linewidth]{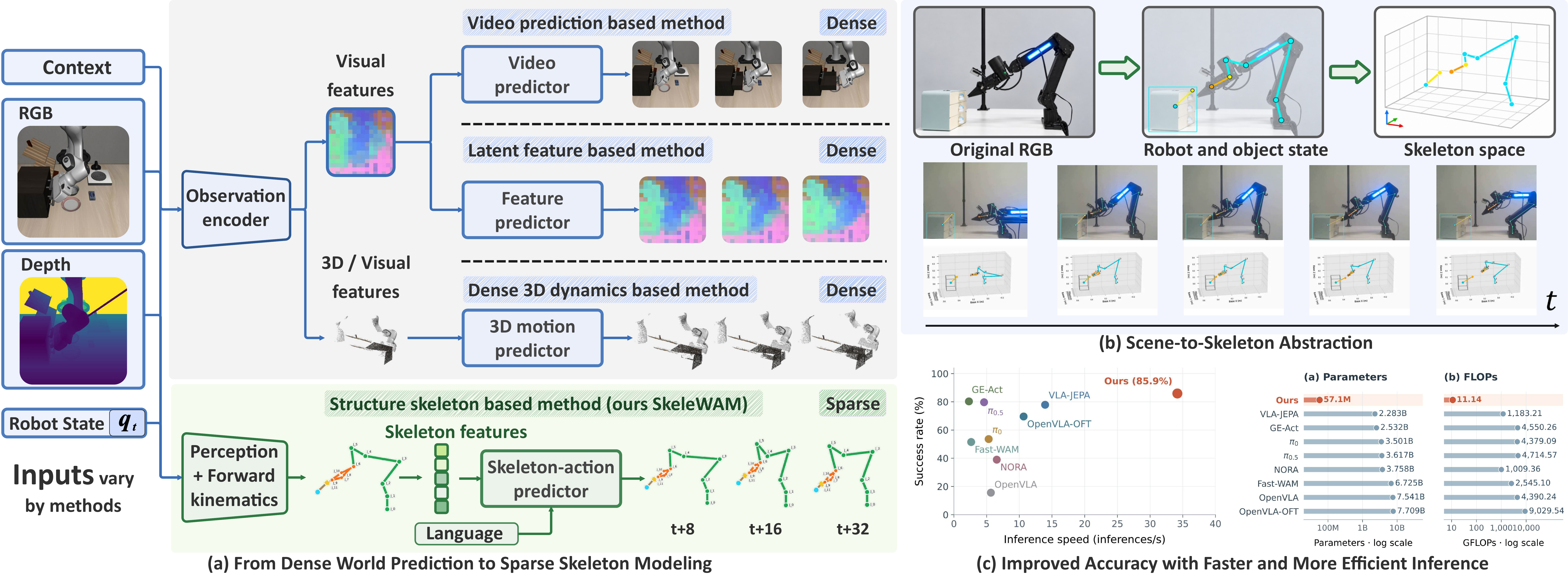}
    \caption{\textbf{Sparse skeleton states enable efficient world action
    modeling.} Existing world action models represent future states as
    videos, visual latents, or dense 3D dynamics. In contrast, SkeleWAM
    converts RGB-D observations and robot states into a sparse 3D skeleton
    comprising robot joints, object centers, and interaction points. During
    training, it jointly learns robot action generation and future skeleton
    prediction. This explicit geometric representation achieves a favorable
    trade-off among task success, inference speed, model size, and
    computational cost.}
    \label{fig:motivation}
\end{figure*}

Building on this representation, we introduce \textbf{SkeleWAM}, a compact world action model that operates on sparse skeleton states. Given current RGB-D observations and robot proprioception, SkeleWAM constructs the skeleton online and encodes it as a set of geometric tokens. During training, the model jointly learns robot action generation and future skeleton prediction, allowing future prediction to supervise the shared geometric representation. At inference, the future prediction branch is omitted, and actions are generated directly from the current skeleton and language instruction. To reduce sensitivity to stochastic sampling, SkeleWAM additionally uses Medoid Action Consensus (MAC) as an auxiliary inference strategy. MAC selects the trajectory with the smallest average distance to other independently sampled candidates, without requiring a reward model or averaging multiple trajectories.

We evaluate SkeleWAM on the full LIBERO-Plus benchmark in a primary observation-based setting, where object landmarks are estimated from RGB-D observations. We additionally report SkeleWAM (sim-state), which replaces these estimates with privileged simulator coordinates solely as a diagnostic reference. In the primary setting, SkeleWAM achieves an overall success rate of 85.9\% with 57.1M parameters, outperforming the strongest compared observation-based baseline, Cosmos-Policy, by 3.7 percentage points. Under camera perturbations, it achieves 93.4\% success, exceeding Cosmos-Policy by 17.6 percentage points. The sim-state variant achieves 87.7\% overall success, leaving a gap of only 1.8 percentage points from the primary setting. These results support our central hypothesis that sparse and explicit robot--object geometry provides an effective state representation for robust manipulation with a compact world action model.

The main contributions of this work are summarized as follows:
\begin{itemize}
    \item We formulate world action modeling in a unified, sparse 3D skeleton state space composed of robot joints, object centers, and interaction points, providing an explicit representation of geometry relevant to manipulation.
    
    \item We introduce SkeleWAM, a compact model that jointly learns robot action generation and future skeleton prediction, with the latter serving as auxiliary supervision during training. SkeleWAM also employs MAC as an auxiliary inference strategy to improve sampling robustness without requiring a reward model.
    
    \item We conduct extensive evaluations on the full LIBERO-Plus benchmark, demonstrating the zero-shot robustness and parameter efficiency of SkeleWAM. Systematic ablations examine the effects of skeleton composition, future skeleton supervision, visual skeleton estimation, and MAC.
    
\end{itemize}

%% file: sec/2_related.tex
\section{Related Work}
\label{sec:related_work}

\subsection{Visuomotor and Vision-Language-Action Models}

Visuomotor policies learn direct mappings from observations to robot actions. ACT predicts temporally coherent action chunks \citep{act}, while Diffusion Policy and DP3 model multimodal action distributions using image and 3D point-cloud observations, respectively \citep{chi2023diffusionpolicy,Ze2024DP3}. Vision-language-action (VLA) models extend this paradigm by incorporating pretrained vision-language representations for language-conditioned multitask control. RT-2 and OpenVLA adapt vision-language representations to action generation \citep{rt2,openvla}, while $\pi_0$ and $\pi_{0.5}$ improve continuous control and generalization through flow matching and heterogeneous co-training \citep{pi0,intelligence2025pi05visionlanguageactionmodelopenworld}. More recent methods improve policy adaptation, action encoding, and model compactness through OpenVLA-OFT, $\pi_0$-FAST, and NORA, respectively \citep{kim2025fine,pertsch2025fast,hung2025nora}. Despite these advances, such policies primarily optimize action prediction and do not explicitly model how manipulation scenes evolve under robot actions.

\subsection{World Action Models}

To capture how manipulation scenes evolve under robot actions, world action models augment action learning with future state prediction. WorldVLA and UniVLA jointly model action generation and visual world evolution \citep{worldvla,wang2026unified}, while DreamZero, GE-Act, and Cosmos-Policy leverage pretrained video models for robot control and planning \citep{dreamzero,geact,kim2026cosmos}. VLA-JEPA replaces pixel reconstruction with the prediction of future latent target representations \citep{vla-jepa}, whereas Fast-WAM retains future video prediction as a co-training objective but removes future generation at inference \citep{fastwam}. Although latent objectives reduce the computational cost and dependence on pixel reconstruction, both videos and visual latents represent interaction geometry only implicitly rather than parameterizing it as explicit state variables. In contrast, SkeleWAM defines both current and future states directly in a sparse 3D skeleton space composed of robot joints, object centers, and interaction points. It jointly learns action generation and future skeleton prediction, allowing future geometric prediction to supervise representation learning during training without being required for action generation at inference.

%% file: sec/3_method.tex
\section{Method}
\label{sec:method}

\subsection{Problem Formulation}

Given a language instruction $\ell$, the current RGB-D observation $o_{t}$, and the robot proprioceptive state $q_{t}$, our objective is to generate an action chunk
\begin{equation}
    \mathbf{A}_{t}
    =
    \left[
        \mathbf{a}_{t},
        \ldots,
        \mathbf{a}_{t+H_a-1}
    \right]
    \in \mathbb{R}^{H_a\times d_a},
\end{equation}
where $\mathbf{a}_{t+i}\in\mathbb{R}^{d_a}$ denotes a single robot action, $H_a$ is the action horizon, and $d_a$ is the action dimension. SkeleWAM represents the current scene as a sparse 3D skeleton,
\begin{equation}
    \mathbf{S}_{t}
    =
    \Phi_{\psi}(o_{t},q_{t},\ell)
    \in \mathbb{R}^{N\times3},
\end{equation}
where $N$ is the number of skeleton nodes and $\Phi_{\psi}$ is the skeleton extractor described in Section~\ref{sec:skeleton_representation}.

For training, we additionally define the future skeleton sequence
\begin{equation}
    \mathbf{S}^{+}_{t}
    =
    \left[
        \mathbf{S}_{t+\delta_1},
        \ldots,
        \mathbf{S}_{t+\delta_{H_s}}
    \right]
    \in \mathbb{R}^{H_s\times N\times3},
\end{equation}
where $H_s$ is the number of future states and
$0<\delta_1<\cdots<\delta_{H_s}$ are their temporal offsets.
Conditioned on $\mathbf{S}_{t}$ and $\ell$, we jointly learn the action and future skeleton distributions,
\begin{equation}
    p^{a}_{\theta}
    \left(\mathbf{A}_{t}\mid\mathbf{S}_{t},\ell\right),
    \qquad
    p^{s}_{\theta}
    \left(\mathbf{S}^{+}_{t}\mid\mathbf{S}_{t},\ell\right),
\end{equation}
where $\theta$ denotes the world action model parameters.
Future skeleton prediction serves as auxiliary supervision during training and is omitted at inference.

\begin{figure*}[t]
    \centering
    \includegraphics[width=1.0\linewidth]{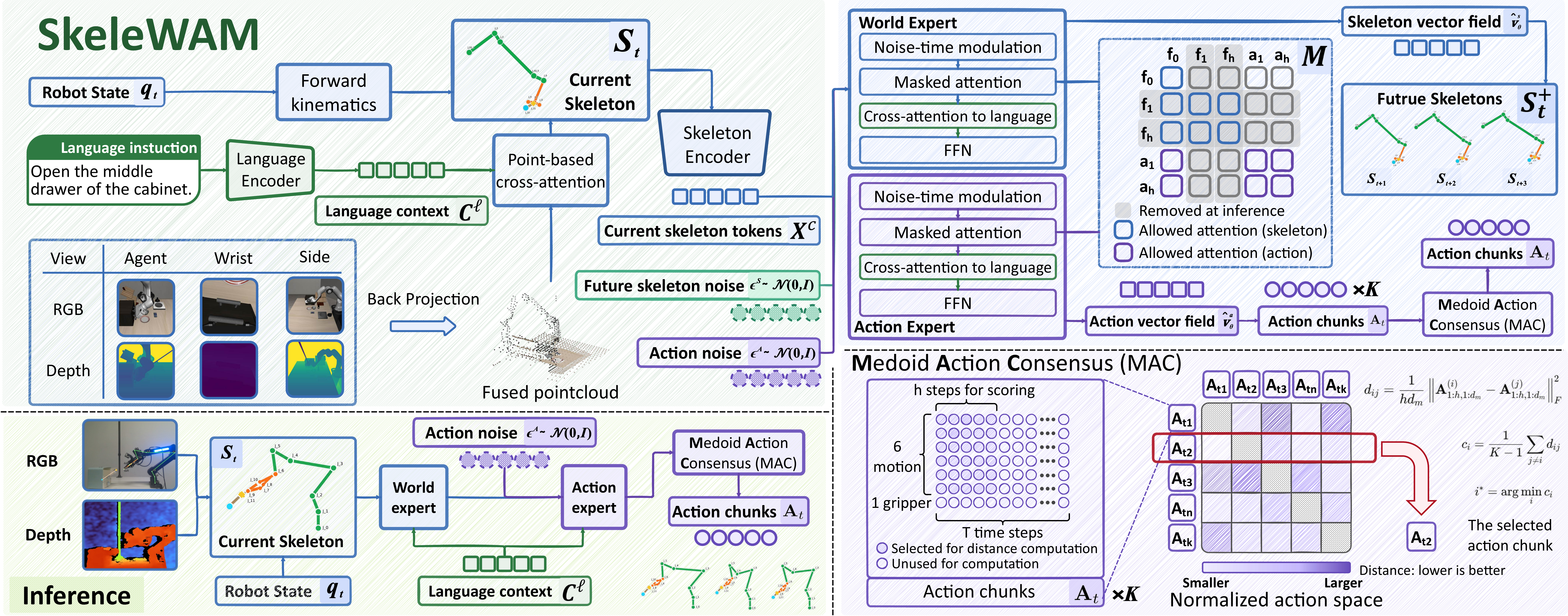}
    \caption{\textbf{Overview of SkeleWAM.}
    Frozen visual perception and forward kinematics construct the current
    skeleton. Action generation and future skeleton prediction are jointly
    trained with shared current-skeleton context. At inference, future
    skeleton prediction is omitted, and Medoid Action Consensus (MAC)
    selects a representative sampled action chunk. The robot executes
    its first $H_e$ actions before observing and replanning.}
    \label{fig:method}
\end{figure*}

\subsection{Skeleton World Representation}
\label{sec:skeleton_representation}

The scene skeleton consists of robot keypoints and task-relevant object landmarks. For object $m$, we define
\begin{equation}
    \mathbf{P}_{t,m}
    =
    \left[
        \mathbf{c}_{t,m};
        \mathbf{u}_{t,m,1};
        \ldots;
        \mathbf{u}_{t,m,L_m}
    \right]
    \in \mathbb{R}^{(1+L_m)\times3},
\end{equation}
where $\mathbf{c}_{t,m}$ denotes the object center and
$\mathbf{u}_{t,m,j}$ denotes its $j$-th interaction point.
The complete skeleton is
\begin{equation}
    \begin{aligned}
        \mathbf{S}_{t}
        &=
        \left[
            \mathbf{J}_{t};
            \mathbf{P}_{t,1};
            \ldots;
            \mathbf{P}_{t,M}
        \right],\\
        N&=N_r+\sum_{m=1}^{M}(1+L_m),
    \end{aligned}
\end{equation}
where $\mathbf{J}_{t}\in\mathbb{R}^{N_r\times3}$ contains robot joint and end-effector positions, $M$ is the number of objects, and semicolons indicate concatenation along the node dimension.

Robot nodes follow the kinematic connectivity, while each object center connects to its associated interaction points. These connections define the skeletal organization; the model predicts node coordinates. Node identities, ordering, and connectivity remain consistent over time.

Robot keypoints are computed through forward kinematics, and object landmarks are estimated from RGB-D observations using a pretrained perception network that remains fixed during training. All coordinates are expressed in a shared robot-centric frame and normalized using training-set statistics. Future targets contain the corresponding nodes at the specified future times in the demonstration trajectories.

\subsection{Skeleton World Action Model}
\label{sec:architecture}

The language instruction is encoded as
$\mathbf{C}^{\ell}=\mathcal{E}_{\mathrm{lang}}(\ell)$.
Separate input adapters map the current skeleton, noisy action chunk, and noisy future skeleton sequence to tokens:
\begin{equation}
    \begin{aligned}
        \mathbf{X}^{c}_{t}
        &=\phi_c(\mathbf{S}_{t}),\\
        \mathbf{X}^{a}_{t,\tau_a}
        &=\phi_a(\widetilde{\mathbf{A}}_{t,\tau_a}),\\
        \mathbf{X}^{s}_{t,\tau_s}
        &=\phi_s(\widetilde{\mathbf{S}}^{+}_{t,\tau_s}),
    \end{aligned}
\end{equation}
where $\tau_a$ and $\tau_s$ denote noise times, distinct from the environment time step $t$. The noisy inputs are defined in Section~\ref{sec:training}; the current skeleton remains unperturbed by generative noise.

Following the two-expert Mixture-of-Transformers design of Fast-WAM \citep{fastwam}, a world expert processes current and future skeleton tokens with shared parameters, while an action expert processes action tokens. Both experts are conditioned on $\mathbf{C}^{\ell}$.

For the three token groups ordered as
$(\mathbf{X}^{c}_{t},\mathbf{X}^{a}_{t,\tau_a},
\mathbf{X}^{s}_{t,\tau_s})$,
we define the block attention mask
\begin{equation}
    \mathbf{M}
    =
    \begin{bmatrix}
        1 & 0 & 0\\
        1 & 1 & 0\\
        1 & 0 & 1
    \end{bmatrix}.
    \label{eq:block_attention_mask}
\end{equation}
Rows index query groups and columns index key and value groups, with $1$ and $0$ indicating allowed and blocked attention. Current skeleton tokens attend within their own group. Each prediction group attends to the current skeleton and itself, with no attention between action and future skeleton tokens. The output heads predict vector fields
$\hat{\mathbf{v}}^{a}_{\theta}\in\mathbb{R}^{H_a\times d_a}$
and
$\hat{\mathbf{v}}^{s}_{\theta}\in\mathbb{R}^{H_s\times N\times3}$
in a single forward pass.

\subsection{Training Objective}
\label{sec:training}

We train both prediction tasks using flow matching. Given a demonstration pair
$(\mathbf{A}_{t},\mathbf{S}^{+}_{t})$,
we sample independent standard Gaussian noise tensors
$\boldsymbol{\epsilon}^{a}$ and $\boldsymbol{\epsilon}^{s}$
with matching shapes, and noise times
$(\tau_a,\tau_s)\sim\rho$ on $[0,1]^2$.
The interpolation paths are
\begin{equation}
    \begin{aligned}
        \widetilde{\mathbf{A}}_{t,\tau_a}
        &=(1-\tau_a)\mathbf{A}_{t}
          +\tau_a\boldsymbol{\epsilon}^{a},\\
        \widetilde{\mathbf{S}}^{+}_{t,\tau_s}
        &=(1-\tau_s)\mathbf{S}^{+}_{t}
          +\tau_s\boldsymbol{\epsilon}^{s},
    \end{aligned}
\end{equation}
where noise time $0$ corresponds to data and $1$ corresponds to Gaussian noise. The associated target vector fields are
\begin{equation}
    \mathbf{v}^{a}
    =\boldsymbol{\epsilon}^{a}-\mathbf{A}_{t},
    \qquad
    \mathbf{v}^{s}
    =\boldsymbol{\epsilon}^{s}-\mathbf{S}^{+}_{t}.
\end{equation}

The joint objective is
\begin{equation}
    \begin{aligned}
        \mathcal{L}_{\mathrm{action}}
        &=\mathbb{E}\left[
            \operatorname{MSE}\left(
                \hat{\mathbf{v}}^{a}_{\theta},
                \mathbf{v}^{a}
            \right)
        \right],\\
        \mathcal{L}_{\mathrm{skeleton}}
        &=\mathbb{E}\left[
            \operatorname{MSE}\left(
                \hat{\mathbf{v}}^{s}_{\theta},
                \mathbf{v}^{s}
            \right)
        \right],\\
        \mathcal{L}
        &=\mathcal{L}_{\mathrm{action}}
          +\lambda_s\mathcal{L}_{\mathrm{skeleton}},
    \end{aligned}
\end{equation}
where $\operatorname{MSE}$ averages over tensor entries and
$\lambda_s$ weights skeleton supervision.
The expectations cover demonstration samples, Gaussian noise, and noise times.
Both losses update the world expert: the skeleton loss supervises future geometric prediction, while the action loss propagates through the current skeleton context.

\begin{figure*}[!t]
    \centering
    \includegraphics[width=\textwidth]{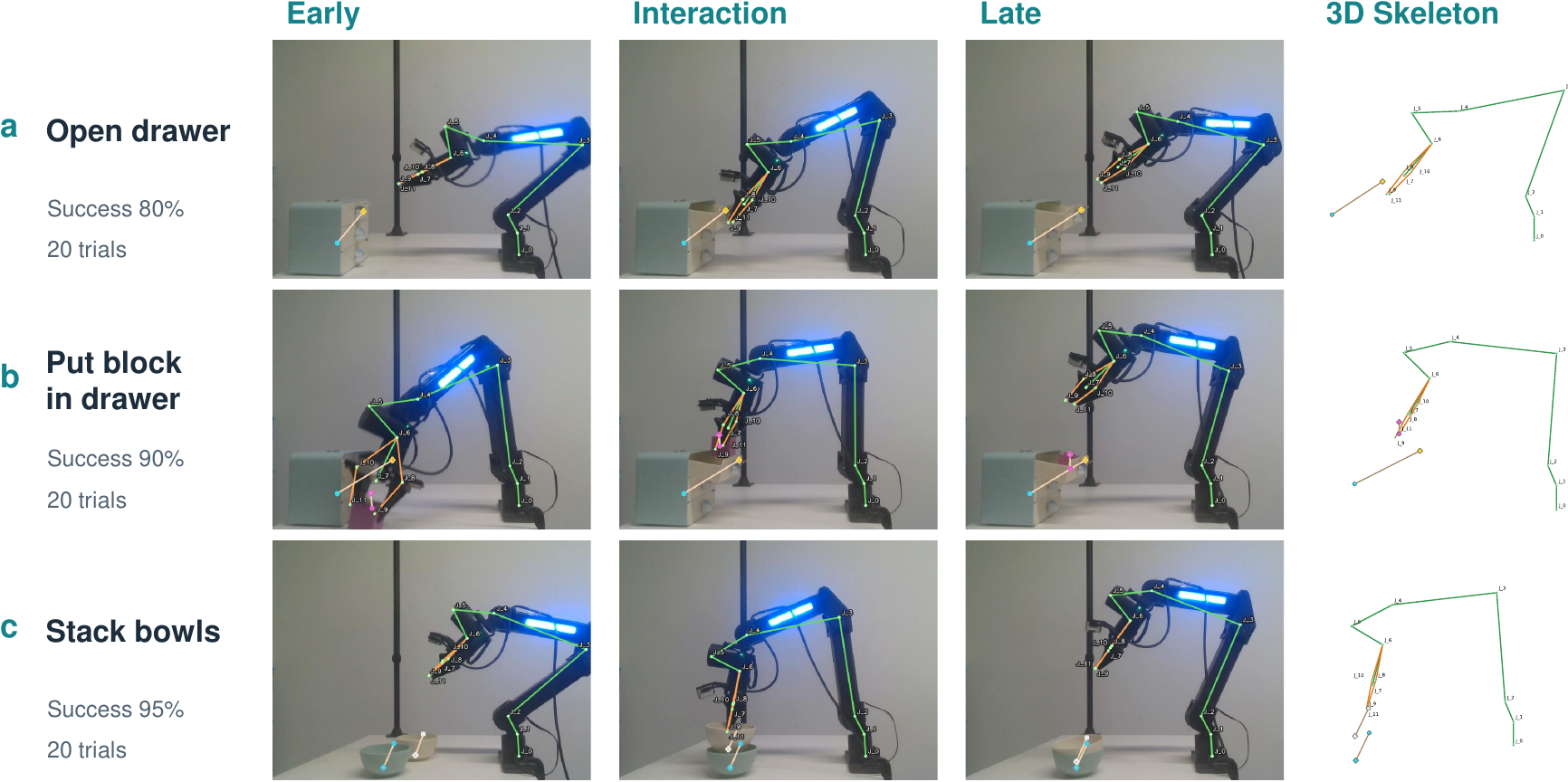}
    \caption{\textbf{Qualitative real-world results.}
    Representative rollouts for opening a drawer, placing a block in a
    drawer, and stacking bowls. Each row shows the early, interaction, and
    late stages of a successful rollout, together with the corresponding
    sparse 3D skeleton. Success rates are computed over 20 trials per task.}
    \label{fig:real_world_qualitative}
\end{figure*}

\subsection{Inference and Medoid Action Consensus}
\label{sec:mac}

At inference, we retain current skeleton processing and the action expert, while omitting future skeleton tokens.
Starting from
$\widetilde{\mathbf{A}}_{t,1}
\sim\mathcal{N}(\mathbf{0},\mathbf{I})$,
we integrate the learned action vector field:
\begin{equation}
    \frac{\mathrm{d}\widetilde{\mathbf{A}}_{t,\tau}}
         {\mathrm{d}\tau}
    =
    \hat{\mathbf{v}}^{a}_{\theta}\left(
        \widetilde{\mathbf{A}}_{t,\tau},\tau;
        \mathbf{S}_{t},\ell
    \right),
    \qquad \tau:1\rightarrow0.
\end{equation}
Here, the vector field is written as a function of the noisy action and current conditions. Integration yields the sampled action chunk
$\mathbf{A}_{t}=\widetilde{\mathbf{A}}_{t,0}$.

We employ Medoid Action Consensus (MAC) as an auxiliary strategy for selecting among stochastic action samples.
Given $K\geq2$ candidates
$\{\mathbf{A}^{(k)}_{t}\}_{k=1}^{K}$
generated from independent noise initializations,
we compare their first $h\leq H_a$ steps over
$d_m$ continuous motion dimensions in normalized action space:
\begin{equation}
    d_{ij}
    =
    \frac{1}{h d_m}
    \left\|
        \bigl(\mathbf{A}^{(i)}_{t}\bigr)_{1:h,1:d_m}
        -
        \bigl(\mathbf{A}^{(j)}_{t}\bigr)_{1:h,1:d_m}
    \right\|_{F}^{2}.
\end{equation}
For notational convenience, the motion dimensions are indexed first.
MAC selects the candidate with the smallest average dissimilarity:
\begin{equation}
    \begin{aligned}
        c_i&=\frac{1}{K-1}\sum_{j\neq i}d_{ij},
        \qquad
        i_{\mathrm{sel}}=\arg\min_i c_i,\\
        \mathbf{A}^{\mathrm{MAC}}_{t}
        &=\mathbf{A}^{(i_{\mathrm{sel}})}_{t}.
    \end{aligned}
\end{equation}
The complete selected chunk is retained, including dimensions excluded from distance computation.
MAC requires neither a reward nor a value model and avoids averaging potentially incompatible trajectories.
For $K=1$, the sole candidate is used directly.
We execute the first $H_e\leq H_a$ actions, update the skeleton from the latest observation, and repeat.
The execution horizon $H_e$ controls replanning, while the consensus window $h$ is used only for candidate selection.

%% file: sec/4_experiment.tex
\begin{table*}[ht]
  \centering
  \caption{Zero-shot success rates (\%) on LIBERO-Plus.
  Parameter counts are approximate.
  Best observation-based results are bolded;
  $^{\S}$ denotes privileged simulator input.}
  \label{tab:libero_plus_wam}
  \setlength{\tabcolsep}{3.4pt}
  \renewcommand{\arraystretch}{1.08}
  \resizebox{\textwidth}{!}{%
  \begin{tabular}{lcrrrrrrrr}
    \toprule
    Method & Params & Camera & Robot & Language & Light &
    Background & Noise & Layout & Overall \\
    \midrule
    OpenVLA \cite{openvla}
      & $\approx 7.5$B & 0.8 & 3.5 & 23.0 & 8.1 &
      34.8 & 15.2 & 28.5 & 15.6 \\
    OpenVLA-OFT \cite{kim2025fine}
      & $\approx 7.7$B & 56.4 & 31.9 & 79.5 & 88.7 &
      93.3 & 75.8 & 74.2 & 69.6 \\
    NORA \cite{hung2025nora}
      & $\approx 3.8$B & 2.2 & 37.0 & 65.1 & 45.7 &
      58.6 & 12.8 & 62.1 & 39.0 \\
    UniVLA \cite{wang2026unified}
      & $\approx 7$B & 1.8 & 46.2 & 69.6 & 69.0 &
      81.0 & 21.2 & 31.9 & 43.9 \\
    $\pi_0$ \cite{pi0}
      & $\approx 3.5$B & 13.8 & 6.0 & 58.8 & 85.0 &
      81.4 & 79.0 & 68.9 & 53.6 \\
    $\pi_0$-Fast \cite{pertsch2025fast}
      & $\approx 3$B & 65.1 & 21.6 & 61.0 & 73.2 &
      73.2 & 74.4 & 68.8 & 61.6 \\
    $\pi_{0.5}$ \cite{intelligence2025pi05visionlanguageactionmodelopenworld}
      & $\approx 3.6$B & 70.6 & 50.5 & 84.4 & 95.7 &
      93.4 & 87.4 & \textbf{84.1} & 79.7 \\
    \midrule
    WorldVLA \cite{worldvla}
      & $\approx 7$B & 0.1 & 27.9 & 41.6 & 43.7 &
      17.1 & 10.9 & 38.0 & 25.0 \\
    Fast-WAM \cite{fastwam}
      & $\approx 6.7$B & 16.4 & 44.5 & 68.9 & 78.2 &
      53.7 & 37.7 & 60.7 & 51.5 \\
    VLA-JEPA \cite{vla-jepa}
      & $\approx 2.3$B & 64.2 & 67.7 & 88.1 & 91.8 &
      93.4 & 65.8 & 83.9 & 77.9 \\
    GE-Act \cite{geact}
      & $\approx 2.5$B & 60.7 & \textbf{77.0} & 77.4 & 95.8 &
      86.0 & 90.9 & 80.2 & 80.3 \\
    Cosmos-Policy \cite{kim2026cosmos}
      & $\approx 2$B & 75.8 & 63.3 & 81.7 & \textbf{96.5} &
      88.9 & 92.7 & 82.2 & 82.2 \\
    \midrule
    \textbf{SkeleWAM (ours)}
      & $\approx 57.1$M & \textbf{93.4} & 71.9 & \textbf{89.1} & 94.7 &
      \textbf{96.0} & \textbf{93.9} & 66.6 & \textbf{85.9} \\
    \midrule
    \textbf{SkeleWAM (sim-state)}$^{\S}$
      & $\approx 51.4$M & 96.1 & 75.2 & 89.0 & 96.6 &
      96.7 & 95.4 & 69.0 & 87.7 \\
    \bottomrule
  \end{tabular}}
\end{table*}

\section{Experiments}
\label{sec:experiments}

\subsection{Simulation Setup}

\paragraph{Benchmark and evaluation.}
We evaluate SkeleWAM on the full LIBERO-Plus benchmark~\citep{libero_plus}, which contains 10,030 variants across seven perturbation categories. We report zero-shot task success rates without fine-tuning on the evaluation variants. Table~\ref{tab:libero_plus_wam} compares SkeleWAM with VLA policies and world-model-based methods.

\paragraph{Baselines.}
We compare SkeleWAM with representative VLA policies, including OpenVLA, OpenVLA-OFT, NORA, UniVLA, $\pi_0$, $\pi_0$-Fast, and $\pi_{0.5}$. We also include methods incorporating world modeling or future-state prediction, including WorldVLA, Fast-WAM, VLA-JEPA, GE-Act, and Cosmos-Policy. SkeleWAM uses a frozen pretrained perception network to estimate object centers and interaction points from RGB-D observations. These object nodes are combined with robot keypoints obtained through forward kinematics to form the current skeleton. SkeleWAM (sim-state) directly obtains the corresponding object coordinates from simulator states and serves as a privileged-input reference.

\paragraph{Implementation details.}
We train and evaluate SkeleWAM on an NVIDIA RTX 4090 GPU. Training runs for 60K steps with an effective batch size of 48, jointly predicting $H_a=32$ action steps and eight future skeleton states. Action and future-skeleton tokens do not attend to each other. At inference, we use 10 flow-integration steps and execute $H_e=16$ actions before replanning. MAC samples $K=3$ action trajectories from independent Gaussian noise initializations and computes consensus over their first $h=10$ motion steps. Future-skeleton prediction is omitted at inference. Additional implementation details are provided in the supplementary material.

\subsection{Simulation Results}

\paragraph{Zero-shot robustness.}
As shown in Table~\ref{tab:libero_plus_wam}, SkeleWAM achieves 85.9\% overall success with 57.1M parameters, outperforming the strongest compared observation-based baseline, Cosmos-Policy (82.2\%), by 3.7 percentage points. Under camera perturbations, SkeleWAM reaches 93.4\%, exceeding Cosmos-Policy by 17.6 points. It also achieves the best observation-based results under language, background, and noise perturbations, with success rates of 89.1\%, 96.0\%, and 93.9\%, respectively. These results support our motivation to model interaction geometry explicitly: representing robot and object states in a shared robot-centric frame provides a compact state space that remains effective under viewpoint and appearance changes.

Performance under layout perturbations remains limited at 66.6\%, compared with 84.1\% for $\pi_{0.5}$. This result suggests that sparse geometric observations alone do not fully address generalization to substantially different spatial arrangements.

\paragraph{Simulator-derived skeletons.}
SkeleWAM (sim-state) achieves 87.7\% overall success with 51.4M parameters, improving over RGB-D by 1.8 percentage points. Its success rates under camera and robot initial-state perturbations increase by 2.7 and 3.3 points, respectively, while language performance remains comparable (89.0\% versus 89.1\%). Since simulator-derived coordinates bypass visual object localization, this setting serves as a privileged-input reference. The relatively small overall gap shows that the frozen perception network provides effective skeleton estimates, while also leaving room for further improvement.

Layout success reaches only 69.0\% even with simulator-derived coordinates. The remaining difficulty under layout perturbations therefore extends beyond visual localization and also reflects limitations in policy generalization to novel spatial configurations.

\subsection{Real-World Experiments}
\label{sec:real_world}

\begin{figure}[t]
    \centering
    \includegraphics[pagebox=cropbox,width=0.95\linewidth]{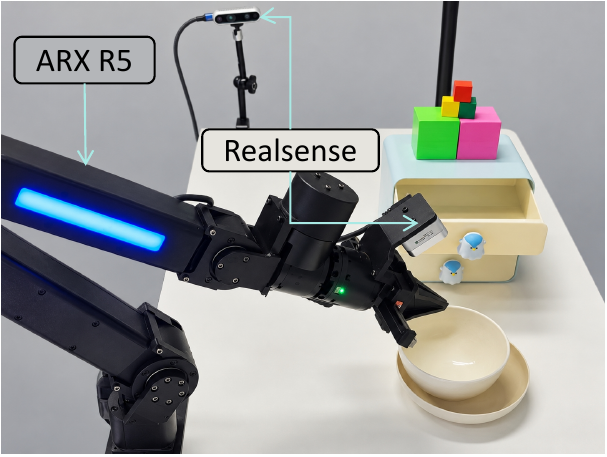}
    \caption{\textbf{Real-world experimental setup.} An ARX R5 robot with external and wrist-mounted RealSense cameras operates in a tabletop workspace containing a drawer unit, blocks, and bowls.}
    \label{fig:setup}
\end{figure}

\paragraph{Experimental setup.}
We evaluate SkeleWAM on an ARX R5 robot with external and wrist-mounted RealSense cameras (Figure~\ref{fig:setup}). We compare against Fast-WAM, $\pi_{0.5}$, and Cosmos-Policy on five real-world tasks, with 20 trials per task for each method. We report per-task success rates and their unweighted mean.

\begin{table*}[t]
    \centering
    \caption{Success rates (\%) on five real-world manipulation tasks.
    Each method is evaluated over 20 trials per task.
    Average is the unweighted mean across tasks.
    Best results in each column, including ties, are bolded.}
    \label{tab:real_world_results}
    \small
    \setlength{\tabcolsep}{5pt}
    \renewcommand{\arraystretch}{1.1}
    \resizebox{\textwidth}{!}{%
    \begin{tabular}{lcccccc}
        \toprule
        Method
        & Open Drawer
        & Close Drawer
        & Stack Blocks
        & Stack Bowls
        & Put Block in Drawer
        & Average $\uparrow$ \\
        \midrule
        Fast-WAM
        & 80 & 85 & 80 & 85 & 85 & 83 \\

        $\pi_{0.5}$
        & 80 & 80 & 70 & 75 & 75 & 76 \\

        Cosmos-Policy
        & \textbf{85} & 85 & \textbf{90} & 90 & 85 & 87 \\

        \midrule
        \textbf{SkeleWAM (ours)}
        & 80 & \textbf{90} & \textbf{90}
        & \textbf{95} & \textbf{90} & \textbf{89} \\
        \bottomrule
    \end{tabular}
    }
\end{table*}

\paragraph{Real-world performance.}
As shown in Table~\ref{tab:real_world_results}, SkeleWAM achieves 89\% average success, exceeding Cosmos-Policy, Fast-WAM, and $\pi_{0.5}$ by 2, 6, and 13 percentage points, respectively. It achieves the highest or joint-highest success rate on four of the five tasks. Figure~\ref{fig:real_world_qualitative} visualizes representative successful rollouts at early, interaction, and late stages. The corresponding 3D skeletons compactly capture the evolving robot--object configurations used for action generation. Together, the quantitative and qualitative results demonstrate the effectiveness of SkeleWAM on the evaluated real-world manipulation tasks.

\subsection{Ablation Studies}
\label{sec:ablations}

We ablate six design choices: object skeleton composition, future-skeleton supervision, cross-branch attention, action execution horizon, the number of MAC candidates, and the consensus window. All experiments use the RGB-D setting on the full LIBERO-Plus benchmark, with fixed training data, optimization settings, and evaluation variants. Unless varied explicitly, we use $H_a=32$, $H_e=16$, $K=3$, and $h=10$. Tables~\ref{tab:ablation_model} and~\ref{tab:ablation_inference} summarize the model and inference ablations, respectively.

\paragraph{Object skeleton composition.}
Keeping robot keypoints unchanged, we compare object representations containing centers only, interaction points only, and both types of points. The combined representation achieves 85.9\% success, outperforming centers only (82.3\%) and interaction points only (77.7\%) by 3.6 and 8.2 percentage points, respectively (Table~\ref{tab:ablation_model}, panel a). Object centers provide stable spatial references, while interaction points describe manipulation-relevant regions. Their combination therefore provides a more complete representation of object interaction geometry.

\paragraph{Contribution of future-skeleton prediction.}
Removing the future-skeleton loss and future tokens reduces success from 85.9\% to 80.1\%, while retaining the same current-skeleton processing and action-generation backbone (Table~\ref{tab:ablation_model}, panel b). Since both variants use the same action-only inference procedure, the 5.8-point improvement comes from future-skeleton supervision during training. This result shows that predicting future geometric states helps the shared world expert learn representations that benefit action generation.

\paragraph{Cross-branch attention.}
We compare future-to-action ($S^+\rightarrow A$), action-to-future ($A\rightarrow S^+$), and no cross-branch attention, while retaining the current-skeleton context and both training objectives. No cross-branch attention achieves 85.9\% success, compared with 85.6\% for $S^+\rightarrow A$ and 84.8\% for $A\rightarrow S^+$ (Table~\ref{tab:ablation_model}, panel c). These relatively small differences, together with the larger drop after removing future-skeleton supervision, indicate that the main benefit comes from co-training through the shared world expert rather than direct information exchange between the two prediction branches. We therefore use no cross-branch attention, which also allows the future-skeleton branch to be removed at inference.

\paragraph{Action execution horizon.}
With $H_a=32$, $K=3$, and $h=10$, executing 5, 16, and 32 actions before replanning yields 80.7\%, 85.9\%, and 81.8\% success, respectively (Table~\ref{tab:ablation_inference}, panel a). The 16-step horizon improves over the shorter and longer settings by 5.2 and 4.1 percentage points. This intermediate horizon balances feedback from updated observations with temporally consistent action execution. We therefore use $H_e=16$ by default.

\paragraph{Number of MAC candidates.}
We vary the number of independently sampled action candidates as $K\in\{1,3,5,10\}$ while fixing $h=10$. Here, $K=1$ corresponds to single-trajectory sampling without MAC. Using three candidates improves success from 84.2\% to 85.9\% (Table~\ref{tab:ablation_inference}, panel b). Increasing $K$ to 5 or 10 yields 85.2\% and 84.1\%, showing that performance does not increase monotonically with the sampling budget. We therefore use $K=3$, which achieves the best result with limited additional inference cost.

\paragraph{Consensus window.}
We vary the trajectory prefix used to compute MAC distances as $h\in\{3,5,10,16\}$ while fixing $K=3$ and $H_e=16$. The corresponding success rates are 85.3\%, 85.4\%, 85.9\%, and 85.7\% (Table~\ref{tab:ablation_inference}, panel c). Performance is relatively stable across the tested windows, with a maximum difference of 0.6 percentage points. The best result is obtained with $h=10$, suggesting that a moderate temporal window captures sufficient trajectory consistency without placing excessive weight on more distant actions. We therefore use $h=10$ by default.

\begin{table}[t]
  \centering
  \caption{Model ablations on the full LIBERO-Plus benchmark
  in the RGB-D setting. Each panel changes one design choice.
  Arrows indicate information flow between action tokens ($A$)
  and future-skeleton tokens ($S^+$). Inference uses
  $H_a=32$, $H_e=16$, $K=3$, and $h=10$.}
  \label{tab:ablation_model}
  \small
  \setlength{\tabcolsep}{8pt}
  \renewcommand{\arraystretch}{1.08}
  \begin{tabular}{lr}
    \toprule
    Configuration & Success (\%) \\
    \midrule
    \multicolumn{2}{l}{\textit{(a) Object skeleton composition}} \\
    Centers only & 82.3 \\
    Interaction points only & 77.7 \\
    Centers + interaction points (default) & \textbf{85.9} \\
    \midrule
    \multicolumn{2}{l}{\textit{(b) Training objective}} \\
    Action only & 80.1 \\
    Action + future skeleton (default) & \textbf{85.9} \\
    \midrule
    \multicolumn{2}{l}{\textit{(c) Cross-branch attention}} \\
    $S^+ \rightarrow A$ & 85.6 \\
    $A \rightarrow S^+$ & 84.8 \\
    No cross-branch attention (default) & \textbf{85.9} \\
    \bottomrule
  \end{tabular}
\end{table}

\begin{table}[t]
  \centering
  \caption{Inference ablations on LIBERO-Plus in the RGB-D setting.
  We fix $H_a=32$. Panel (a) varies $H_e$ with $K=3$ and $h=10$;
  panel (b) varies $K$ with $H_e=16$ and $h=10$;
  panel (c) varies $h$ with $H_e=16$ and $K=3$.}
  \label{tab:ablation_inference}
  \small
  \setlength{\tabcolsep}{8pt}
  \renewcommand{\arraystretch}{1.08}
  \begin{tabular}{lr}
    \toprule
    Configuration & Success (\%) \\
    \midrule
    \multicolumn{2}{l}{\textit{(a) Action execution horizon}} \\
    $H_e=5$ & 80.7 \\
    $H_e=16$ (default) & \textbf{85.9} \\
    $H_e=32$ & 81.8 \\
    \midrule
    \multicolumn{2}{l}{\textit{(b) Number of MAC candidates}} \\
    $K=1$ (without MAC) & 84.2 \\
    $K=3$ (default) & \textbf{85.9} \\
    $K=5$ & 85.2 \\
    $K=10$ & 84.1 \\
    \midrule
    \multicolumn{2}{l}{\textit{(c) Consensus window}} \\
    $h=3$ & 85.3 \\
    $h=5$ & 85.4 \\
    $h=10$ (default) & \textbf{85.9} \\
    $h=16$ & 85.7 \\
    \bottomrule
  \end{tabular}
\end{table}

%% file: sec/5_conclusion.tex
\section{Conclusion}

We introduced SkeleWAM, a compact world action model that represents
manipulation scenes as sparse 3D skeletons composed of robot joints, object
centers, and interaction points. By jointly learning robot action generation
and future skeleton prediction, SkeleWAM uses future geometric dynamics as
auxiliary supervision during training without requiring visual reconstruction
or additional world prediction at inference. We further introduced Medoid
Action Consensus as a lightweight strategy for selecting a representative
trajectory from stochastic action samples without relying on a reward model.

Experiments on the full LIBERO-Plus benchmark and five real-world manipulation
tasks demonstrate that explicit robot--object geometry provides an effective
and parameter-efficient state space for world action learning. SkeleWAM
achieves strong zero-shot robustness with substantially fewer parameters than
the compared VLA and world-model-based methods, while the ablations confirm the
benefits of combining object centers with interaction points and of predicting
future skeleton states during training. The remaining performance gap under
large layout changes suggests that sparse geometry alone does not fully capture
the variation required for spatial generalization. Future work may address
this limitation by incorporating uncertainty-aware perception, adaptive
interaction landmarks, and richer relational structure while preserving the
compactness of the skeleton representation.